\documentclass[times,review,10pt]{elsarticle}
\usepackage{amssymb}
\usepackage{amsmath}
\usepackage{graphicx}
\usepackage{hyperref}
\usepackage{algorithm}
\usepackage{algorithmic}
\usepackage{amssymb}
\usepackage{amsmath}
\usepackage{booktabs}

\journal{}

\begin{document}

\begin{frontmatter}



\title{ECAFormer: Low-light Image Enhancement using Dual Cross Attention}


% \author{
%     %Authors
%     % All authors must be in the same font size and format.
%     Yudi Ruan\textsuperscript{\rm 1},
%     Hao Ma\textsuperscript{\rm 2},
%     Weikai Li\textsuperscript{\rm 1}\thanks{Corresponding Authors},
%     Xiao Wang\textsuperscript{\rm 2*}\\
%    % Francisco Cruz\equalcontrib,
%    % Marc Pujol-Gonzalez\equalcontrib
% }
\author[1]{Yudi Ruan}
\author[2]{Hao Ma}
\author[3]{Di Ma}
\author[1]{Weikai Li\corref{cor1}}
\author[2]{Xiao Wang\corref{cor1}}

\cortext[cor1]{Corresponding author}
\affiliation[1]{organization={School of Mathematics and Statistics, Chongqing Jiaotong University},%Department and Organization
            city={Chongqing},
            postcode={400074}, 
            % state={Chongqing},
            country={China}}
\affiliation[2]{organization={School of Artificial Intelligence, Anhui University},%Department and Organization
            city={Hefei},
            postcode={230039}, 
            state={AnHui},
            country={China}}
\affiliation[3]{organization={College of Information Science and Technology \& Artificial Intelligence, Nanjing Forestry University},%Department and Organization
            city={Nanjing},
            postcode={211106}, 
            state={Jiangsu},
            country={China}}

%% Abstract
\begin{abstract}
Low-light image enhancement (LLIE) aims to improve the perceptibility and interpretability of images captured in poorly illuminated environments. Existing LLIE methods often fail to capture the non-local self-similarity and long-range dependencies, causing the loss of complementary information between multiple modules or network layers, ultimately resulting in the loss of image details. To alleviate this issue, we design a hierarchical mutual Enhancement via a dual cross-attention transformer (ECAFormer), which introduces an architecture that enables concurrent propagation and interaction of multiple disentangling features. To capture the non-local self-similarity, we design a Dual Multi-head self-attention (DMSA), which leverages the disentangled visual and semantic features across different scales, allowing them to guide and complement each other. Further, a cross-scale DMSA block is incorporated to capture residual connections, thereby integrating cross-layer information and capturing the long-range dependencies. Experimental results show that the ECAFormer reaches competitive performance across multiple benchmarks, yielding nearly a 3.7\% improvement in PSNR over the suboptimal method, demonstrating the effectiveness of information interaction in LLIE. For facilitating the efforts to replicate our results, our implementation is available on GitHub \footnote{\href{https://github.com/ruanyudi/ECAFormer}{https://github.com/ruanyudi/ECAFormer}.}.
\end{abstract}


% \begin{highlights}
% \end{highlights}

%% Keywords

\end{frontmatter}




\section{Introduction}
\label{sec:intro}

Capturing images in low-light conditions presents significant challenges in computer vision, including the degradation of fine details, reduced color saturation, diminished contrast and dynamic range, and uneven exposure. These issues can greatly compromise the integrity and clarity of captured data, negatively impacting the effectiveness of subsequent applications such as autonomous driving systems \cite{ranft2016role, cai2020vtgnet, bresson2017simultaneous} and nighttime surveillance efforts \cite{fu2021let, BULUSWAR1998245}. Naturally, enhancing the visibility of objects and details in low-light images is crucial, possessing far-reaching implications for a spectrum of vision applications.
%目前基于深度学习的方法大致可以分为以下三个类别：Conventional Method、Guided Method、Two-Branch Method。其中，传统的方法主要依赖于数据驱动和损失函数的驱动。因此，传统的方法受限于数据的复杂性，模型难以学习到较好的表达方式。后来，研究逐渐转向于通过基于先验人为知识构建合适的网络。Guided Method通常基于人为知识将输入分解为需要保持的本质成分和需要修改的目标成分。该方法在目标成分通过特定网络来增强的同时，利用本质成分来引导，以此来尽可能维持本质成分。例如，在增强的过程中，应当在修改光照分量的同时，尽可能地保持反射分量的一致性。与之类似的是Two-Branch的方法，此类方法将输入分解为相对可分的子成分，并分别通过特定的增强网络进行修改。 总的来说，在构建的模型的时候加入人为知识能够有效引导模型的学习。但是，这两类方法通常忽略了输入的不同子成分之间，在潜在空间中的联系。由于受限于模型架构，现有的方法无法很好地学习到不同成分之间的潜在联系。
%为了解决LLIE领域中现有的问题，we proposed a Hierarchical Mutual Enhancement via a Cross-Attention transformer (ECAFormer) as shown in Figure 1. 

\begin{figure}[t]
  \centering
  \includegraphics[width=0.8\textwidth]{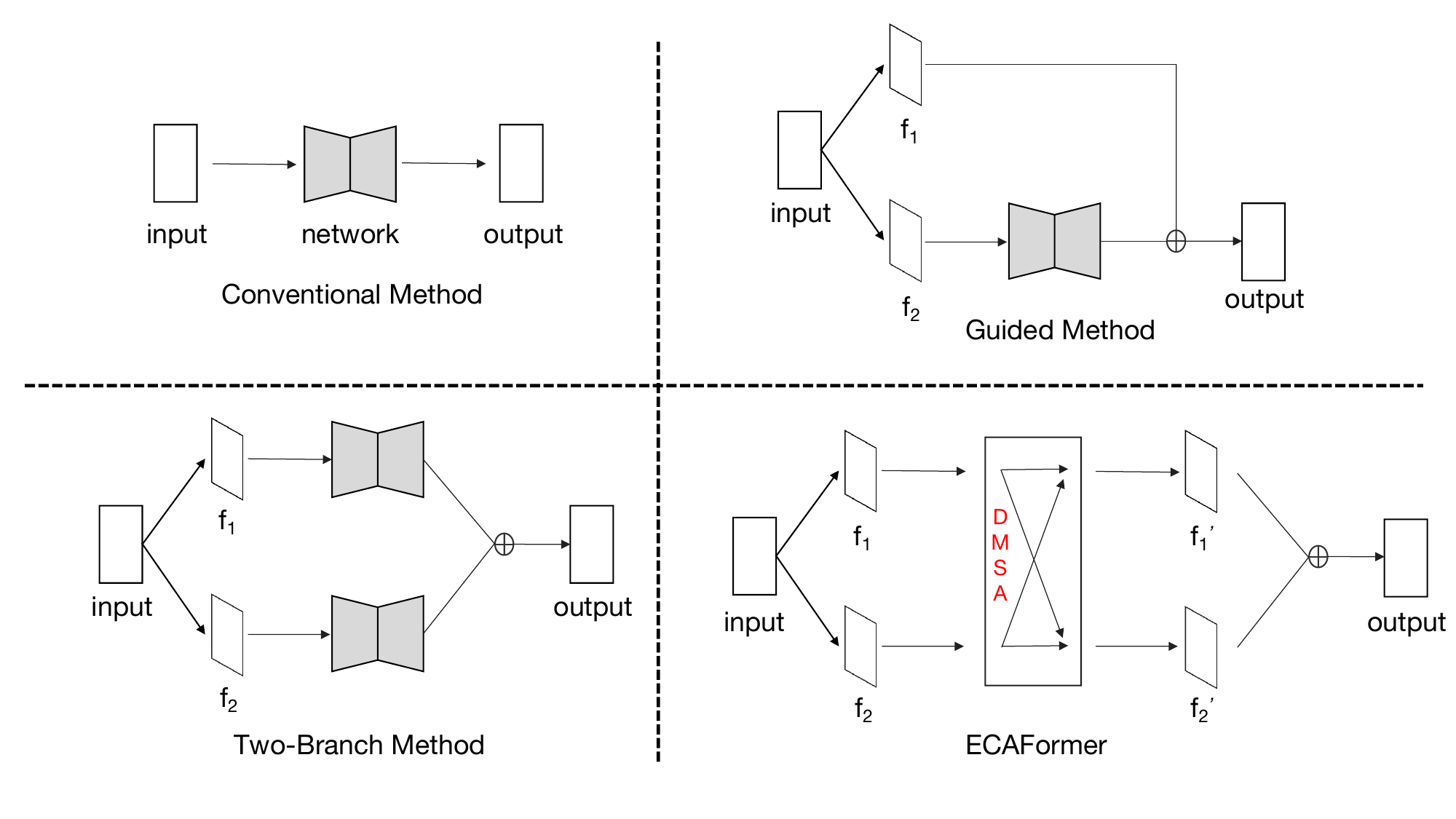}
  \caption{Main architecture comparison. Compared to other methods, our approach utilizes DMSA to simultaneously propagate two features forward and facilitate interactions at different scales. This method is advantageous for extracting latent connections between features.
  \label{fig:archcompar}
  }
  %主要架构对比: 与其他方法相比，我们的方法通过DMSA能够同时前向传播两个特征，并在不同尺度下做交互。这种方法利于提取特征间的潜在联系。
\end{figure}

Currently, efforts for LLIE have predominantly utilized deep-learning techniques, which can be categorized into three main approaches: Conventional, Guided, and Two-Branch Methods, as depicted in Figure \ref{fig:archcompar}. Conventional methods, such as LLNet \cite{lore2017llnet}  and UFormer \cite{wang2022uformer}, hinge on data-driven strategies and tailored loss functions but are constrained by data complexity, hindering the model's ability to capture robust representations. To overcome this, guided methods \cite{ZHANG2023106972, ZHANG2024107793} integrate prior knowledge into LLIE. They decompose the input into essential and target components, leveraging the former knowledge to guide the enhancement of the latter through specialized networks. For example, some guided methods \cite{ZHUANG2021104171, LI2023106457} based on Retinex theory \cite{land1971lightness} focus on preserving the consistency of the reflection component while adjusting the illumination. Another method for incorporating prior knowledge is the Two-Branch Methods, which involve refining the input by dividing it into separate sub-components and processing each component with its dedicated enhancement network \cite{ma2021learning,tang2022drlie, AHERRAHROU2023109643, HE2023106969}. However, despite the use of prior knowledge, these methods fail to capture the non-local self-similarity and long-range dependencies. This limitation, inherent in the model architecture, can result in a loss of image details during enhancement.

Regarding the current situation in the LLIE field, we focus on enhancing the information interactions across different disentangled components and layers to fill in the image details. To achieve this, we employ the U-shaped architecture as the foundation of our model to enhance low-light images. Specifically, we proposed ECAFormer as shown in Figure  \ref{fig:MainFrame}. Following the idea of two branch methods, we incorporate a knowledge-driven strategy that disentangles the images into distinct visual and semantic components, which facilitates detail preserving with the guidance of semantic information. To capture non-local similarities, we emphasize leveraging the interactive advantages of both visual and semantic components by integrating a modified transformer that effectively harnesses these elements. Specifically, to enhance the interactions across different sub-components in the latent space, we designed DMSA to handle multiple inputs, which captured potential relationships between different components. Notably, the transition from Multi-Head Self-Attention (MHSA) in the vanilla transformer to DMSA modules is strategically designed to simultaneously exploit both long- and short-range dependencies within the image data. Furthermore, to capture long-range dependencies, we designed a Cross-Scale DMSA that facilitates the interaction between residual and current layer information. This mechanism is ingeniously leveraged to refine the residual connections within ECAFormer. Our main contribution can be summarized as follows:

%在微光图像中，不同的区域可以具有不同的亮度、噪声、可见性等特征。亮度极低的区域会受到噪声的严重破坏，而同一图像中的其他区域仍然可以有合理的能见度和对比度。为了更好的整体图像增强，我们应该自适应地考虑弱光图像中的不同区域。因此，我们通过探索信噪比(signal-To-noise- ratio, SNR)[3, 54]来研究图像空间中信噪比的关系，以实现空间变化增强。1. 低信噪比的区域通常是不清楚的。因此，我们在更长的空间范围内利用非局部图像信息进行图像增强。 (Long-range)2. 信噪比较高的区域通常具有较高的能见度和较少的噪声。因此，本地图像信息通常就足够了。(Short-Range)
\begin{itemize}

\item We explore a novel LLIE method named ECAFormer, which illustrates the importance of information integration across layers and components for detail preservation in LLIE.
\item 
We collect a novel \textit{Traffic-297} dataset, which includes various traffic
scenes and provides a new benchmark in LLIE.%design a novel DMSA module instead of vanilla MHSA to balance both visual and semantic components. The addition of DMSA enables the network to learn the latent relationships between different sub-components.
\item We achieve competitive results on six LLIE datasets, emphasizing computational frugality and parameter parsimony.
\end{itemize}
The paper is structured as follows. In Section \ref{Relatedwork}, we respectively reviewed the related work in the field of low light enhancement and cross-attention mechanisms. In Section \ref{Methodology}, we introduced the problem definition, the proposed method, and the details of model optimization. In Section \ref{Experiment}, we compared our model with other methods using six publicly available datasets and conducted ablation experiments to validate its effectiveness. Finally, we summarized in Section \ref{Conclusion} and proposed directions for future work.

% The following paper is organized as follows. In Section \ref{Relatedwork}, we respectively reviewed the related work in the field of low light enhancement and cross-attention mechanisms. In Section \ref{Methodology}, we introduced the problem definition, the proposed method, and the details of model optimization. In Section \ref{Experiment}, we compared our model with seven publicly available datasets and conducted ablation experiments to validate its effectiveness. Finally, we summarized in Section \ref{Conclusion} and proposed directions for future work.

\section{Related Work}
\label{Relatedwork}
\subsection{Convolutional Neural Networks based LLIE}
%low light image enhancement在计算机视觉任务中尤为重要。在极其缺乏光照的条件下获取的图像将大大降低计算机视觉任务的性能。传统的方法大致可以分为基于直方图和基于Retinex理论两个类别。
%LLIE is particularly important in computer vision tasks. Traditional LLIE methods can be broadly categorized into histogram-based and Retinex-based approaches. For example, CLAHE \cite{pizer1990contrast} normalizes the value of each pixel in an image based on a histogram function. BPDHE \cite{ibrahim2007brightness} extends traditional histogram equalization by producing an output image whose mean intensity closely matches that of the input.
%The Retinex model \cite{land1977retinex} and its multi-scale variant \cite{jobson1997multiscale} decompose brightness into illumination and reflectance components, which are then processed independently. These manually crafted constraints and priors lack sufficient self-adaptivity to accurately recover image details and colors, often leading to the obliteration of details, local under- or over-saturation, uneven exposure, or the emergence of halo artifacts around objects. 

The rapid advancement in low-light image dataset collection has led to the emergence of numerous deep learning-based enhancement methods \cite{li2021low, LI2025109749}. LLNET \cite{lore2017llnet} introduced a variant of the stacked sparse denoising autoencoder for improving degraded images, establishing a foundational framework for the application of deep learning in image enhancement. RetinexNet \cite{wei2018deep} employed a deep Retinex-based architecture to enhance low-light images by decomposing them into illumination and reflectance components. RUAS \cite{liu2021retinex} utilized an advanced search unfolding technique based on a Retinex architecture. EnlightenGAN \cite{jiang2021enlightengan} innovatively used a generative inverse network as the primary framework, initially training with unpaired images. LEDNet \cite{zhou2022lednet}, is a robust network specifically designed to address the dual challenges of low-light enhancement and deblurring simultaneously. LACN \cite{FAN2023105632} proposed a novel module by introducing the attention mechanism into the ConvNeXt backbone network.

However, these Convolutional Neural Networks (CNN) based models fail to capture long-range dependencies, resulting in the loss of detailed information. To alleviate this, this paper introduces the transformer architecture and specifically designed cross-scale DMSA block, which significantly enhances the model's ability to extract long-range dependencies and global information.

\subsection{Disentangling based LLIE}
%Representation decomposition aims to model data with diverse variations by decomposing complex data into more manageable sub-components \cite{zhang2022learning}. In recent years, this concept has gradually been adopted in the LLIE field. 

Compared to conventional LLIE methods \cite{LIU2024109012, ZHU2023106866, CHEN2024109207}, the disentangling-based techniques have more effectively captured the latent knowledge within the data and significantly enhanced model performance. Among these approaches, it can be classified into two types, as illustrated in Figure \ref{fig:archcompar}: (1) Two-Branch Methods, which optimize distinct features using separate networks, such as CSDGAN \cite{ma2021learning} and DRLIE \cite{tang2022drlie}; (2) Guided Methods, which use one feature to guide the optimization of another feature, such as SNRNet \cite{xu2022snr} and RetinexFormer \cite{cai2023retinexformer}.

However, the aforementioned model architectures inadequately exploit the complementary information among the decomposed sub-components, frequently addressing them in isolation. This approach hinders the network's ability to learn the non-local self-similarity and the intrinsic relationships between multiple sub-components. Therefore, we propose the DMSA module, which can simultaneously handle and inter-relate two features during forward computation. This modification enables the model to learn the relationships between sub-components, integrating their respective advantages, thus effectively capturing the non-local self-similarity and the intrinsic relationships to improve the final output of the model.
%特征分解的目的是对变化多样的数据进行建模，将复杂的数据分解为利于处理的子成分。近些年来，低光增强领域也渐渐采用这种思想。各种由理论知识辅助的分解策略的引入，相比于传统的方法，能更好地学到数据中的潜在知识，显著提高了模型的性能。在这些方法中，特征的组合框架可以分为（1）Two-branch Method： 这类方法通常通过两个网络分别对两个特征做优化。例如：CSDGAN通过IENet推断光照特征，RENet推断反射特征,合理地利用了物理理论; DRLIE将图像解藕成scene content和exposure attribution，并分别用网络进行优化。（2）Guided Method：这类方法将其中一个特征作为引导来优化另一个特征。例如：SNRNet先将图像分解出SNR Map，用于辅助后续损失函数和特征融合。However，上述提到的两种模型架构无法全面的利用分解后的子成分，在计算往往使子成分过于独立，无法使网络学习到多个子成分之间的内在联系。因此我们提出来一个能够在前向计算时同时处理两个特征并互相辅助的DMSA模块。这一更改能够让模型参与子成分间关系的学习，能够综合其各自的优点，抑制理想假设的偏见，从而使模型最终的输出达到更好的结果。

\begin{figure}[t]
    \centering
    \includegraphics[width=\textwidth]{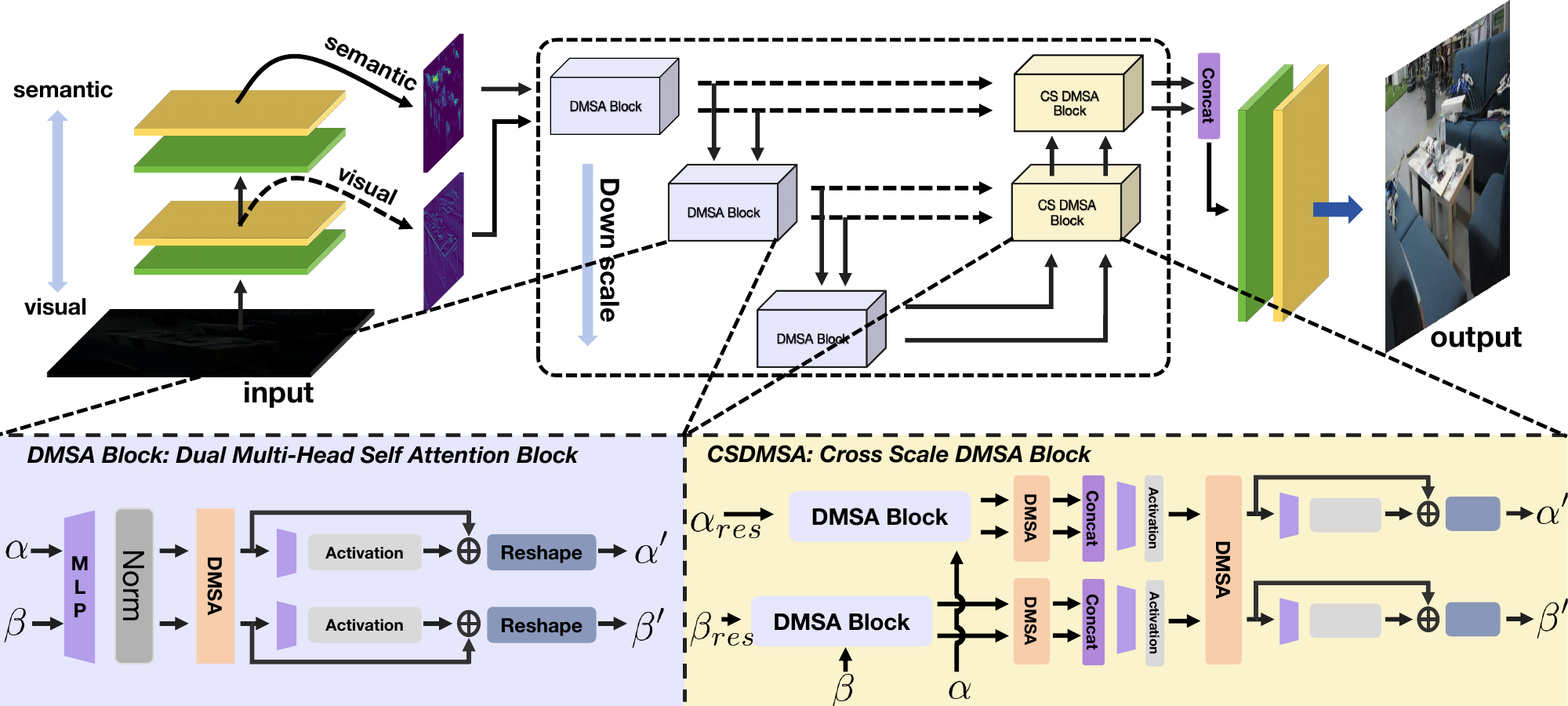}
    \caption{The flowchart of ECAformer mainly consists of three parts: (1) Visual-Semantic Convolution modules \ref{shallow-deep}, which output short-range features (visual-features) and long-range features (semantic-features). (2)The U-shaped cross-attention Transformer \ref{mutual-guidance} engages with long- and short-term input features through DMSA, concurrently propagating these features. (3) Mapping convolution, where the module projects the interacted features back to image features.${\in\mathbb{R}^{\emph{C}\times\emph{H}\times\emph{W}}}$.}
    \label{fig:MainFrame}
\end{figure}

\section{Methodology}
\label{Methodology}
In this section, we present a comprehensive description of each network module proposed and the corresponding loss functions employed. Inspired by the successful U-shaped architecture in LLIE \cite{li2021low}, we introduce a transformer network with a U-shaped architecture, depicted in Figure \ref{fig:MainFrame}. Specifically, ECAFormer comprises three main components: (1) Visual-semantic convolution module \ref{shallow-deep}, which generates short-range features and long-range features. (2) The U-shaped cross attention Transformer \ref{mutual-guidance} interacts with long- and short-term input features through DMSA, simultaneously propagating these features. (3) Mapping convolution, where the module projects the interacted features back to image features ${\in\mathbb{R}^{\emph{C}\times\emph{H}\times\emph{W}}}$. In practice, the low-light input image undergoes convolution filtering to extract semantic information and visual detail information respectively. The enhanced image is then obtained through interaction enhancement of a U-shaped network using semantic and visual features. %The network model is optimized through VGG perceptual loss and Charbonnier  Loss.
% \subsection{Problem definition}
% The problem definition of low light image enhancement is mainly to deal with the problems of low brightness, low contrast, noise, and artifacts in images with insufficient lighting. These problems are mainly caused by the loss of visual information due to insufficient light or complex environments when images are taken in low light environments. Therefore, the task of low light intensity enhancement is to restore and enhance the damaged image information through algorithms and technical means, so that it can be better recognized and understood by human visual systems. The problem can be modeled as Eq. \ref{modelEquation}
% \begin{equation}
% \label{modelEquation}
%     \hat{I}=f(I,\theta)
% \end{equation}
% where ${\emph{I}\in\mathbb{R}^{\emph{C}\times\emph{H}\times\emph{W}}}$ denotes the input images, ${{\hat{I}}\in\mathbb{R}^{\emph{C}\times\emph{H}\times\emph{W}}}$ denotes the enhanced images, and ${\theta}$ denotes the parameters of the network  \emph{f}. The target of the optimization step is finding a suitable parameter ${\theta}$ to minimize differences between high-light images and enhanced images. It can be defined as Eq. \ref{modelTarget}. 
% \begin{equation}
% \label{modelTarget}
%     \hat{\theta}=\mathop{\arg\min}\limits_{\theta}(\ell(\hat{I},R))
% \end{equation}
% where $\emph{R}$ denotes well-exposed images, $\ell$ denotes the loss function which is used to guide the network.
\subsection{Problem Definition}
Low-light image enhancement aims to mitigate the challenges associated with low brightness, low contrast, noise, and artifacts in images captured under inadequate lighting conditions. These issues arise from the loss of visual information under insufficient illumination or complex environmental factors at the time of image capture. The objective of low-light image enhancement is to utilize models to restore and enhance low-light images $\mathbb{L}$, bringing them closer to normal images $\mathbb{N}$ and rendering them more perceptible to the human visual system.

The problem can be mathematically modeled as follows:
\begin{equation}
\label{modelEquation}
    \hat{I} = f(I, \theta),
\end{equation}
where ${I \in \mathbb{R}^{C \times H \times W}}$ represents the input low-light images, ${\hat{I} \in \mathbb{R}^{C \times H \times W}}$ represents the enhanced images, and ${\theta}$ represents the parameters of the network function $f$. 

The optimization objective is to find the optimal parameters ${\theta}$ that minimize the differences between the enhanced images and the reference high-light or normal images. This can be formulated as:
\begin{equation}
\label{modelTarget}
    \hat{\theta} = \mathop{\arg\min}\limits_{\theta} \ell(\hat{I}, R),
\end{equation}
where ${R \in \mathbb{R}^{C \times H \times W}}$ denotes the well-exposed reference or normal images, and ${\ell}$ denotes the loss function used to guide the network.

\subsection{Model Framework}
\subsubsection{Preliminary}
In LLIE, CNNs are adept at extracting intricate local features, while Transformer networks excel in capturing valuable global feature information from complex environments. As the depth of the convolution layers increases in CNNs, the extracted features progressively embody richer semantic information. Consequently, features obtained from initial shallow convolution operations are predominantly visual, while those derived from deeper convolution layers contain more sophisticated semantic insights. To capitalize on these characteristics, we have developed a decoupling convolution extractor, which decouples the image to the visual and semantic features.

\begin{equation}
\begin{aligned}
\label{conv feature}
    &f_{v} = Conv_1(I), \\
    &f_{s} = Conv_2(f_{v}), 
\end{aligned}
\end{equation}
where $f_{v}$ retains an abundance of detailed visual features, while $f_{s}$ preserves advanced semantic features. Both attributes are pivotal for achieving the enhanced final result.

The attention mechanism facilitates interactivity among elements, significantly enhancing global feature extraction capabilities. The conventional self-attention mechanism is defined by Eq. (\ref{self-attention}), where the vectors $Q$, $K$, and $V$ represent query, key, and value derived from a single input. This operation selectively focuses on crucial information, optimizing resource utilization and swiftly capturing the most relevant data. Leveraging this advantage, we have developed a unique DMSA to facilitate the fusion of two distinct features. This module processes two inputs and enables their interaction via the attention mechanism while preserving their dimensional integrity. A detailed discussion of this module will be provided later.
%in section \ref{2.2.3}.
\begin{equation}
\label{self-attention}
    \text{MHSA}(Q,K,V) = \text{softmax}\left(\frac{QK^T}{\sqrt{d_k}}\right)V.
\end{equation}
We have developed a U-shaped network structure with enhanced residual information, specifically tailored to improve the network's capacity for multi-scale interaction. This architecture facilitates a comprehensive synthesis of features across various scales. During the down-sampling phase, we strategically employed a 2$\times$ scaling factor to compel the network to undergo a rigorous compression process, as we separate the image into visual and semantic features. This intentional compression is crucial as it allows the network to extract more refined global information, which is essential for understanding broader contextual cues. The methodology and details of the down-sampling stage are illustrated in Eq. (\ref{downsampleStage}).
\begin{equation}
\label{downsampleStage}
[f_{v}, f_{s}]^{(i)} = 
    \text{RS}(\text{DMSA}_{i}([
    f_{v},f_{s}]^{(i-1)})),    
\end{equation}
where $\{i\in{1, 2\}}$ denotes the step of the down-sample stage, $RS(\cdot)$ represents the resample operation, the $DMSA_{i}$ denotes the DMSA Block varies at $i$ stages. At the bottom of the U-shaped network, we used a bottleneck consisting of two DMSA modules. And in the up-sampling process, we employed operations proportional to down-sampling. 
During the up-sampling phase, we utilized CSDMSA to effectively retain and restore intricate details that are often lost during the down-sampling process through residual information. The residual information initially engages in cross attention with their corresponding features, followed by cross attention between the two types of features in the CSDMSA block. This process is crucial in maintaining the fidelity of feature representations, ensuring that the reconstructed outputs closely resemble the original inputs. Finally, we obtain the final output through concatenation and mapping convolution. It can be represented by Eq. (\ref{finalOutput}).
\begin{equation}
\label{finalOutput}
\hat{I} = \mathcal{A}_{agg} \left( f_{v}^{(T)}, f_{s}^{(T)} \right).
\end{equation}

\subsubsection{Visual-Semantic Convolution Module}
\label{shallow-deep}
\begin{figure}[t]
    \centering
    \includegraphics[width=0.8\textwidth]{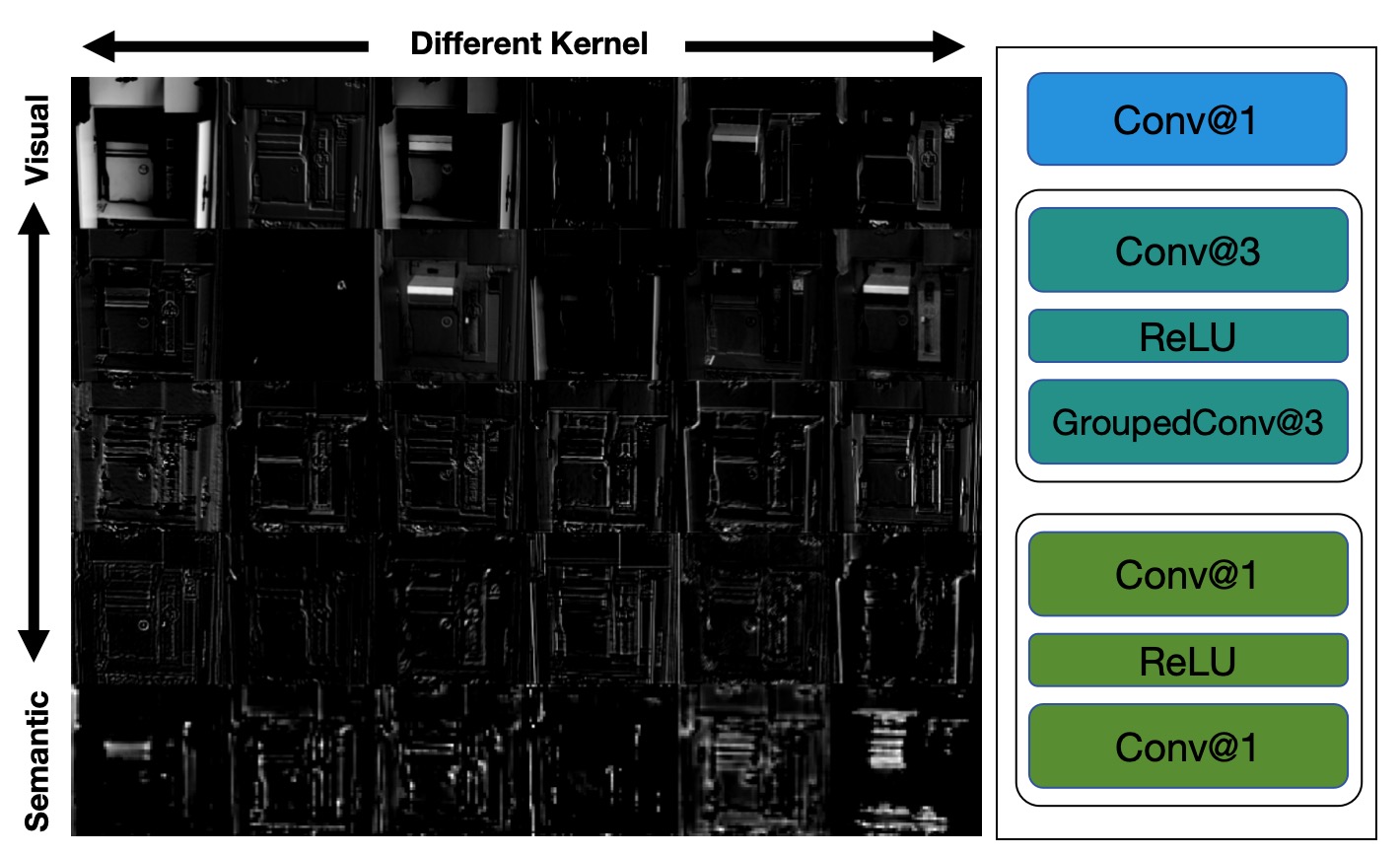}
    \caption{ From top to bottom are the shallow outputs and deep outputs of the model, respectively. }
    \label{fig:enter-label2}
\end{figure}

Inspired by the potent capability of convolution layers to enhance local features, we devised a visual-semantic convolution module specifically designed to capture local characteristics. In CNNs, as the number of convolution layers increases, the receptive field of the model progressively enlarges, leading to the extraction of increasingly complex semantic features as shown in Figure \ref{fig:enter-label2}. It can be observed that visual feature ${f_{v}}$ focuses more on fine details, while semantic feature ${f_{s}}$ emphasizes broader connections and contextual relationships within the image. However, this expansion often results in the attenuation of fine-grained detail within the features. To address this, we introduce a dedicated convolution module that outputs two distinct types of features: ${f_{v}}$ and ${f_{s}}$. Here, $f_{v}$ is derived from shallow convolution layers, capturing detailed visual features, whereas $f_{s}$ emanates from deeper convolution layers, encapsulating higher-level semantic information. Additionally, the CNNs can capture periodic and local spatial features, addressing the spatial context induction bias in transformers that rely solely on positional embedding. We employ depth-wise separable convolution within this module to enhance the rate of forward propagation without compromising much accuracy. Upon processing through this module, the network then employs attention mechanisms to facilitate a dynamic interaction between these two distinct features.

\subsubsection{DMSA}
\label{mutual-guidance}
\begin{figure}[t]
    \centering
    \includegraphics[width=\textwidth]{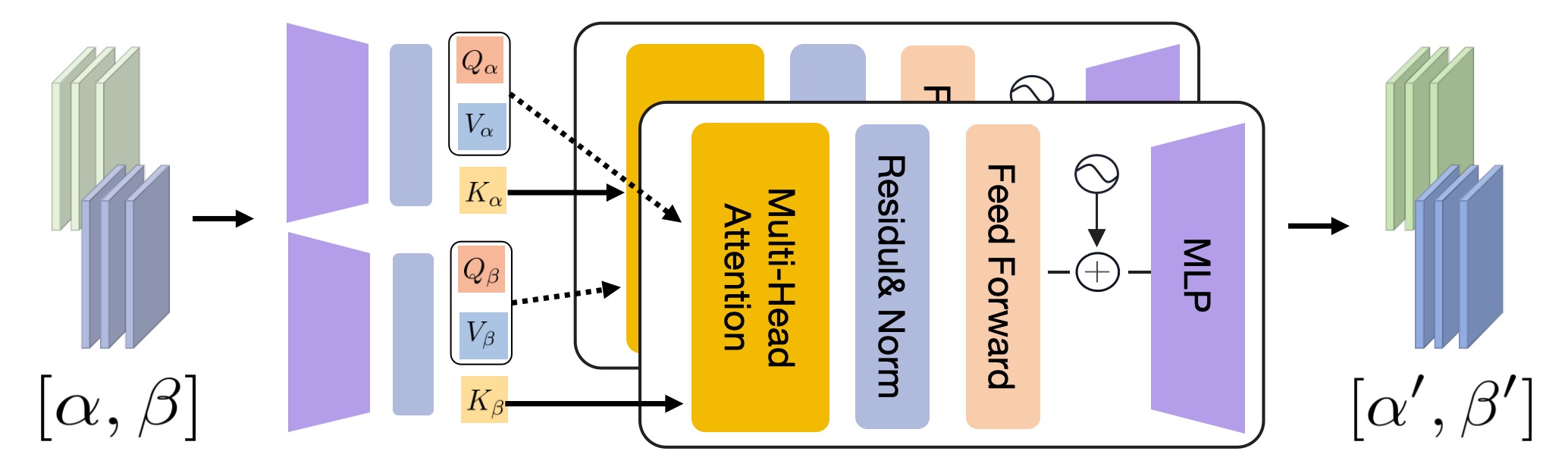}
    \caption{DMSA:  A highly symmetrical module is depicted in the figure, which illustrates the process of computing the $[\alpha',\beta']$ guided by each other using cross attention.}
    \label{fig:enter-label}
\end{figure}
\label{2.2.3}
Self-attention is highly efficient at directing concentrated attention toward pivotal information, thereby optimizing resource utilization and rapidly acquiring the most relevant data. However, its standard configuration is restricted to processing only a single input, which can limit its applicability in scenarios requiring complex interactions between multiple data streams. To address this constraint, we have innovated the DMSA module as shown in Figure \ref{fig:enter-label}, a sophisticated mechanism designed to effectively manage and integrate dual inputs. In this module, inputs $\alpha$ and $\beta$ are processed to generate distinct sets of $Q$, $K$, and $V$ vectors through separate mappings. We strategically cross the key vectors of both inputs within this module to enhance their inter-connectivity, fostering a richer, more comprehensive interaction. This is achieved by the multiplication of the query vector with the corresponding crossed key vector as in Eq. (\ref{MGMSA_definition}), producing an attention map that encapsulates the intricate dynamics between the two inputs. Enhanced by the adaptive sampling operations of the U-shaped network, this setup facilitates a profound interaction across multiple scales, allowing for a nuanced integration of features that is essential for analysis and interpretation tasks.
\begin{equation}
\label{MGMSA_definition}
\begin{aligned}
    \text{DMSA}(Q_\alpha,K_\beta,V_\alpha) =  \text{softmax}(Q_\alpha K_\beta^T*\zeta)V_\alpha.
\end{aligned}
\end{equation}
The scaling factor ${\zeta}$ is determined through optimization. To preserve the spatial positional relationships among pixels, we have incorporated a position embedding ($PosEmb$) module, executed via convolution operations. Consequently, the output of the Dual Multi-head Self Attention module $[\alpha', \beta']$ can be formally defined in Eq. (\ref{whole output}).
\begin{equation}
\label{whole output}
\begin{aligned}
    [\alpha', \beta']=[&\text{DMSA}(Q_\alpha,K_\beta,V_\alpha)+\text{PosEmb}(V_\alpha), \\
    &\text{DMSA}(Q_\beta,K_\alpha,V_\beta)+\text{PosEmb} (V_\beta)].
\end{aligned}
\end{equation}

%在下采样过程中，往往会导致细节的丢失从而导致图像模糊，因此残差信息的利用至关重要。因此我们采用了特殊设计的DMSA模块来如何跨层的残差信息，来减少由于下采样导致的图像模糊。
% \subsubsection{CSDMSA}
% During the downsampling process, detail loss often leads to image blurring, making utilizing residual information crucial. Therefore, we employ a specially designed DMSA module to handle cross-layer residual information, aiming to reduce image degradation.
% \begin{equation}
% \begin{aligned}
%     & [\alpha_{res}',\alpha_{mid}]=DMSA([\alpha_{res},\alpha]), \\
%     & [\beta_{res}',\beta_{mid}]=DMSA([\beta_{res},\beta]).
% \end{aligned}
% \end{equation}
% Then, we concatenated the corresponding features to facilitate subsequent interaction operations.
% \begin{equation}
% \begin{aligned}
%     &\alpha_{agg}=W*concat([\alpha_{res}',\alpha_{mid}])+B,\\
%     &\beta_{agg}=W*concat([\beta_{res}',\beta_{mid}])+B,
% \end{aligned}
% \end{equation}
% where $W$ is the learned weight matrix, $B$ is the learned bias matrix.Finally, the interaction between $\alpha_{agg}$ and $\beta_{agg}$ is processed through a DMSA module, followed by a resampling operation. The final output of CSDMSA is as follows Eq.\ref{csdmsaoutput}.
% \begin{equation}
% \begin{aligned}
%     \label{csdmsaoutput}
%     &[\alpha',\beta']\\
%     &=CSDMSA([\alpha,\beta])\\
%     &=RS(DMSA([\alpha_{agg},\beta_{agg}])).
% \end{aligned}
% \end{equation}
\subsubsection{CSDMSA}

During the downsampling process, detail loss often leads to image blurring, making the utilization of residual information crucial. To address this, we employ a specially designed Cross-Scale Dual Multi-head Self Attention (CSDMSA) module to handle cross-layer residual information, aiming to reduce image degradation.

First, the residual and intermediate features are processed by the DMSA module as follows:
\begin{equation}
\begin{aligned}
    & [\alpha_{res}',\alpha_{mid}'] = \text{DMSA}([\alpha_{res}, \alpha_{mid}]), \\
    & [\beta_{res}',\beta_{mid}'] = \text{DMSA}([\beta_{res}, \beta_{mid}]),
\end{aligned}
\end{equation}
%\alpha_{mid},\beta_{mid}代表的是当前层的特征，\alpha_{res},\beta_{res}分别为对应的残差信息.
where \(\alpha_{mid}\) and \(\beta_{mid}\) represent the features of the current layer, while \(\alpha_{res}\) and \(\beta_{res}\) correspond to the residual information, respectively.
Next, we concatenate the corresponding features to facilitate subsequent interaction operations:
\begin{equation}
\begin{aligned}
    & \alpha_{agg} = W \cdot \text{concat}([\alpha_{res}', \alpha_{mid}]) + B, \\
    & \beta_{agg} = W \cdot \text{concat}([\beta_{res}', \beta_{mid}]) + B,
\end{aligned}
\end{equation}
where \(\alpha_{agg}\) and \(\beta_{agg}\) represent the features after fusion, \(W\) is the learned weight matrix and \(B\) is the learned bias matrix.
Finally, the interaction between \(\alpha_{agg}\) and \(\beta_{agg}\) is processed through another DMSA module, followed by a resampling operation. The final output $[\alpha', \beta']$ of the CSDMSA module is given by:
\begin{equation}
\begin{aligned}
    \label{csdmsaoutput}
     [\alpha', \beta'] 
    &= \text{CSDMSA}([\alpha, \beta]) \\
    &= \text{RS}(\text{DMSA}([\alpha_{agg}, \beta_{agg}])),
\end{aligned}
\end{equation}
where $\text{RS}(\cdot)$ denotes the resampling operation.

\subsection{Loss Function}
We incorporated two types of loss functions that better align with human visual perception and facilitate faster model training. The Total Loss $\mathcal{L}_{Total}$ can be represented as Eq. (\ref{totalloss}), $\lambda\in{[0,1]}$.
\begin{equation}
\label{totalloss}
    \mathcal{L}_{Total} = \lambda*\mathcal{L}_p + (1-\lambda)*\mathcal{L}_c,
\end{equation}
where $\mathcal{L}_p$ 
 is the loss of perceptual and $\mathcal{L}_c$ is the loss of  charbonnier.
\subsubsection{Perceptual Loss}
\label{perceptualloss}
Perceptual Loss \cite{johnson2016perceptual} adopts an efficient approach by quantifying the discrepancies through the squared error between features extracted from specific layers or an aggregation of multiple layers after both the ground truth and the reconstructed image have traversed the same pre-trained neural network. 
% This technique transcends traditional pixel-based difference calculations, offering a superior method for handling anomalies and enhancing the overall robustness of the model. By focusing the loss calculations on the variances within the deeper layers of feature representations, Perceptual Loss adeptly captures and emphasizes high-level semantic nuances. Consequently, this approach produces images that more accurately reflect human visual perceptions, thereby aligning the output more closely with the intricacies of human sight. 
The Perceptual Loss $\mathcal{L}_p$ is given as follows:
\begin{equation}
\label{perceptualLossFrom}
    \mathcal{L}_p = \sum_{i=1}^{n} \frac{1}{C_i H_i W_i} \left| F_{i}^{l}(f(I)) - F_{i}^{l}(R) \right|^2.
\end{equation}
Among them, \emph{f} symbolizes the enhancement network. ${F_{i}^{l}}$ indicates the i-th feature map in the $l$-th layer. We have utilized a VGG-19 network pre-trained on ImageNet and employed the output feature maps from its initial five ReLU layers to compute the loss. This method leverages the deep network's architecture to extract rich, complex feature representations that are critical for assessing the perceptual quality of the enhanced images.

\subsubsection{Charbonnier Loss}
\label{cbloss}
Compared to the conventional $L_{1}$ Loss, Charbonnier Loss exhibits superior robustness and stability, particularly when dealing with outliers. Additionally, the computational efficiency of Charbonnier Loss is enhanced due to its reliance on a single square and root operation, in contrast to the ordinary $L_{1}$ Loss which necessitates an absolute value computation. This streamlined calculation speeds up the processing and contributes to smoother gradients, facilitating more effective optimization during model training. The Charbonnier Loss $\mathcal{L}_c$ is given as follows:
\begin{equation}
    \mathcal{L}_c = \sum_{i=1}^{n} \sqrt{\left(\hat{I} - R\right)^2 + \epsilon^2}.
\end{equation}
Benefiting from the addition of $\epsilon$, the phenomenon of gradient vanishing when $\hat{I}$ and \emph{R} are very close has been alleviated, making the model easier to train.

\begin{figure}[t]
\centering
\includegraphics[width=\textwidth]{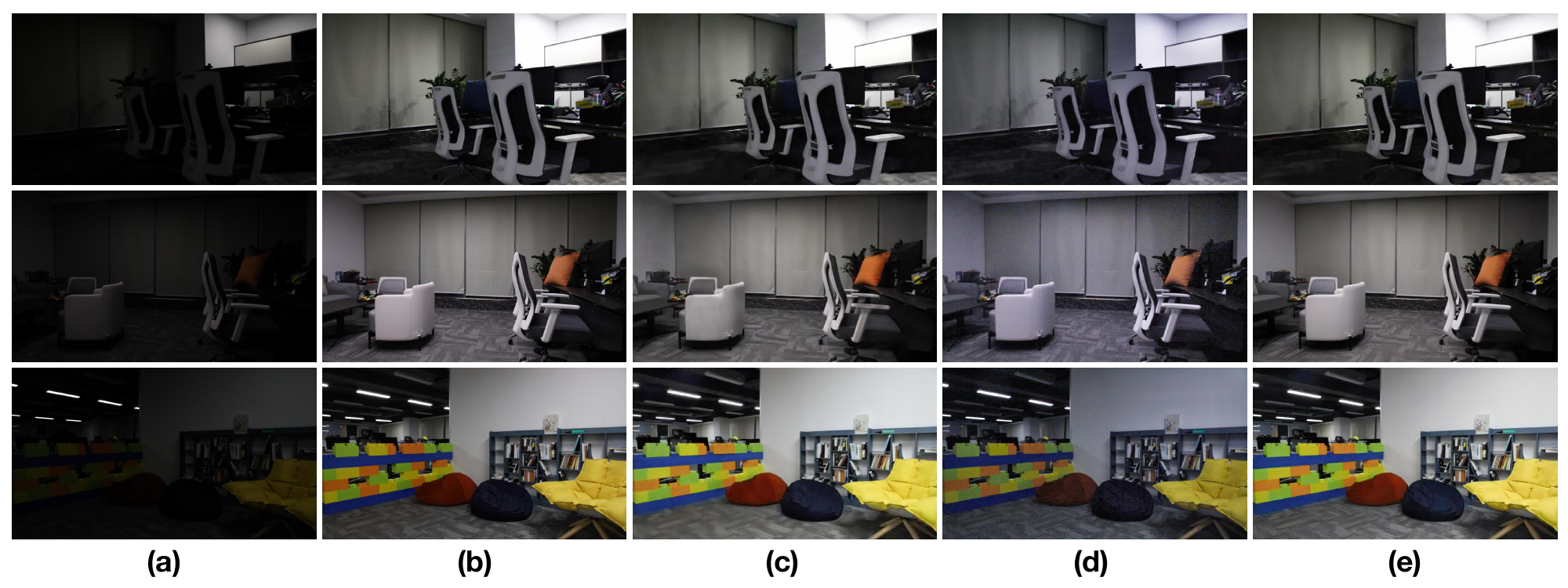}
\caption{Three images from SDSD-indoor test set were selected for comparison with different methods. (a) Input Image. (b) Ground Truth. (c) ZeroDCE++. (d) SNRNet. (e) ECAFormer. }
\label{fig:visualResults0}
\end{figure}
%From the images, it can be observed that our method produces results that appear more natural.

\begin{table}[]
\centering
\caption{Comparison of PSNR of different methods across 5 public datasets. The data was obtained either by training with publicly available code or from the information provided in the paper. ECAFormer achieved promising results without a significant increase in parameters.}
% \hspace{}
\resizebox{\textwidth}{!}{
\begin{tabular}{cccccc|cc}
\toprule
Methods                  & LOL-v1         & LOLv2-r         & LOLv2-s         & SID              & SMID             & AVG         & Params           \\
\midrule
RN(BMVC 2018)            & 16.77          & 15.47            & 17.13            & 16.48            & 22.83            & 17.74            & 0.84             \\
KinD(ACMMM 2019)         & 20.86          & 14.74            & 13.29            & 18.02            & 22.18            & 17.82            & 8.02             \\
EnGAN(TIP 2021)          & 17.48          & 18.23            & 16.57            & 17.23            & 22.62            & 18.43            & 114.35           \\
RUAS(CVPR 2021)          & 18.23          & 18.37            & 16.55            & 18.44            & 25.88            & 19.49            & 0.00             \\
DRBN(TIP 2021)           & 20.13          & 20.29            & 23.22            & 19.02            & 26.60            & 21.85            & 5.27             \\
UF(CVPR 2022)            & 16.36          & 18.82            & 19.66            & 18.54            & 27.20            & 20.12            & 5.29             \\
RSTM(CVPR 2022)          & 22.43          & 19.94            & 21.41            & 22.27            & 26.97            & 22.60            & 26.13            \\
MIRNet(TPAMI 2022)       & 24.14          & 20.02            & 21.94            & 20.84            & 25.66            & 22.52            & 31.76            \\
SNR(CVPR 2022)           & \textbf{24.61} & \underline{21.48}      & 24.14            & 22.87            & \underline{28.49}      & \underline{24.32}      & 4.01             \\
LLF(AAAI 2023)           & 23.65          & 20.06            & 24.04            & \textbackslash{} & \textbackslash{} & \textbackslash{} & 24.55            \\
RF(ICCV 2023) & 23.69          & 21.43            & \underline{24.16}      & \underline{23.85}      & 27.55            & 24.14            & 1.61             \\
U2E-Net(PR 2024)         & 23.92          & \textbackslash{} & \textbackslash{} & \textbackslash{} & \textbackslash{} & \textbackslash{} & \textbackslash{} \\
\midrule
ECAFormer                & \underline{24.24}    & \textbf{21.99}   & \textbf{25.86}   & \textbf{24.69}   & \textbf{29.34}   & \textbf{25.22}   & 2.50           \\
\bottomrule
\end{tabular}
}
\label{psnrcomparison}
\end{table}
\section{Experiment}
\label{Experiment}

\begin{figure}[]
    \centering
    \includegraphics[width=\textwidth]{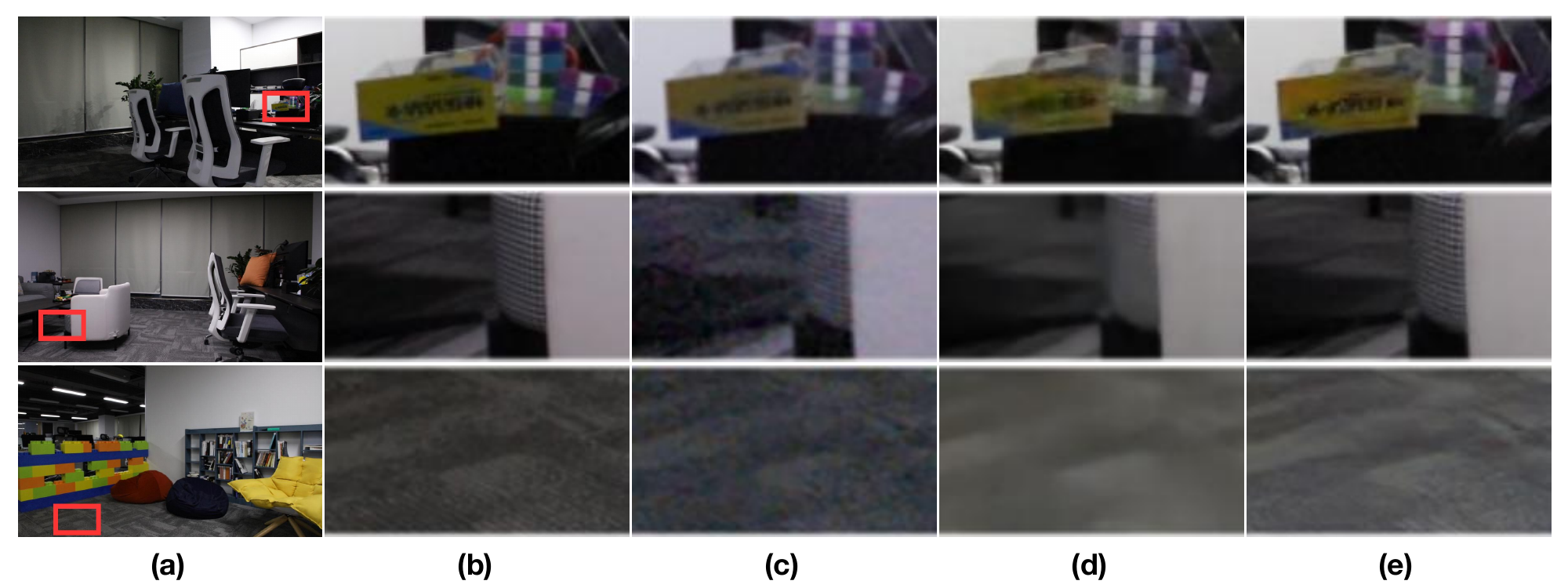}
    \caption{The detailed comparison of the three images selected from SDSD-indoor test set. (a) Indicator. (b) Ground Truth. (c) ZeroDCE++. (d) SNRNet. (e) ECAFormer.  }
    \label{fig:visualResults1}
\end{figure}
\begin{figure}[]
    \centering
    \includegraphics[width=\textwidth]{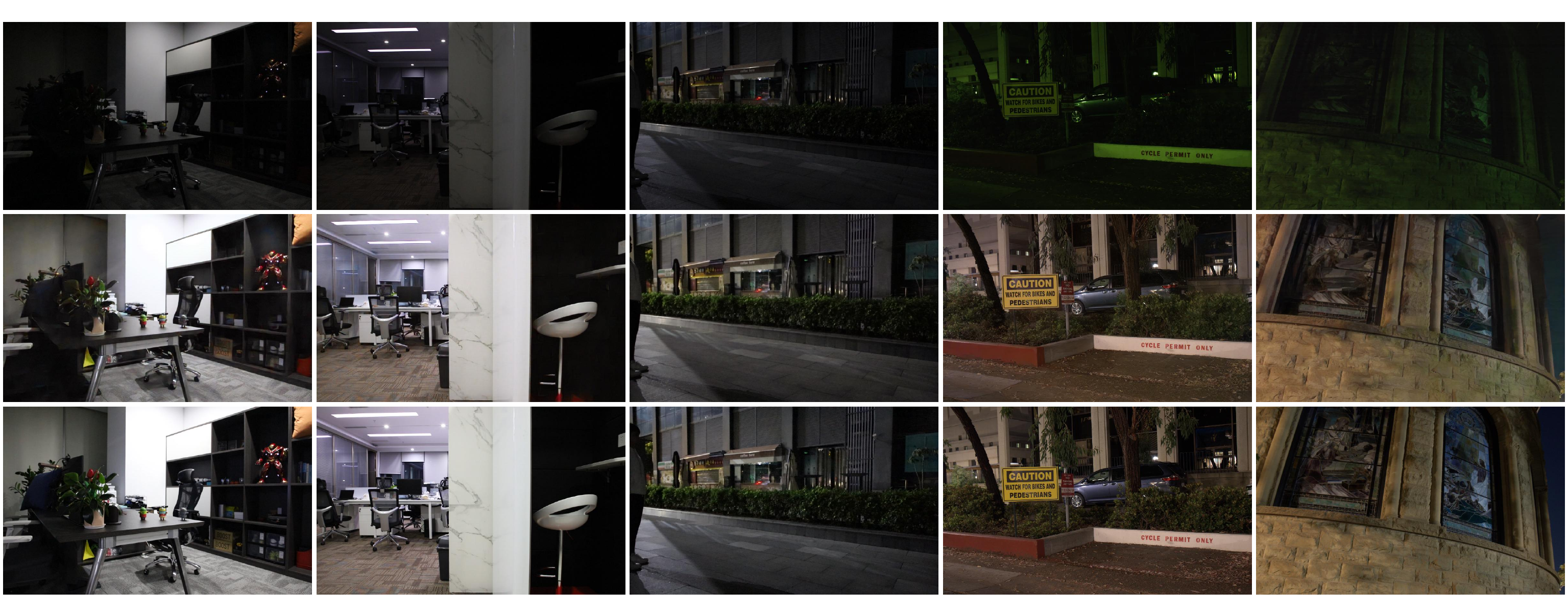}
    \caption{ The experimental results across different scenes. From the first row to the third row are (1) Input Image. (2) ECAFormer. (3) Ground Truth. }
    % 我们在多个复杂场景中测试了我们的方法，结果表明ECAFormer具有良好的表现。这些图片来自于SDSD-indoor、SDSD-outdoor、SMID。
    \label{fig:visualResults3}
\end{figure}
%It can be seen that our method effectively retains details while suppressing noise.

%We tested our method in complex scenarios, and the results indicate that ECAFormer performs well. These images are from SDSD-in, SDSD-out, and SMID datasets.
%htbp
\begin{figure}[t]
    \centering
    \includegraphics[width=\textwidth]{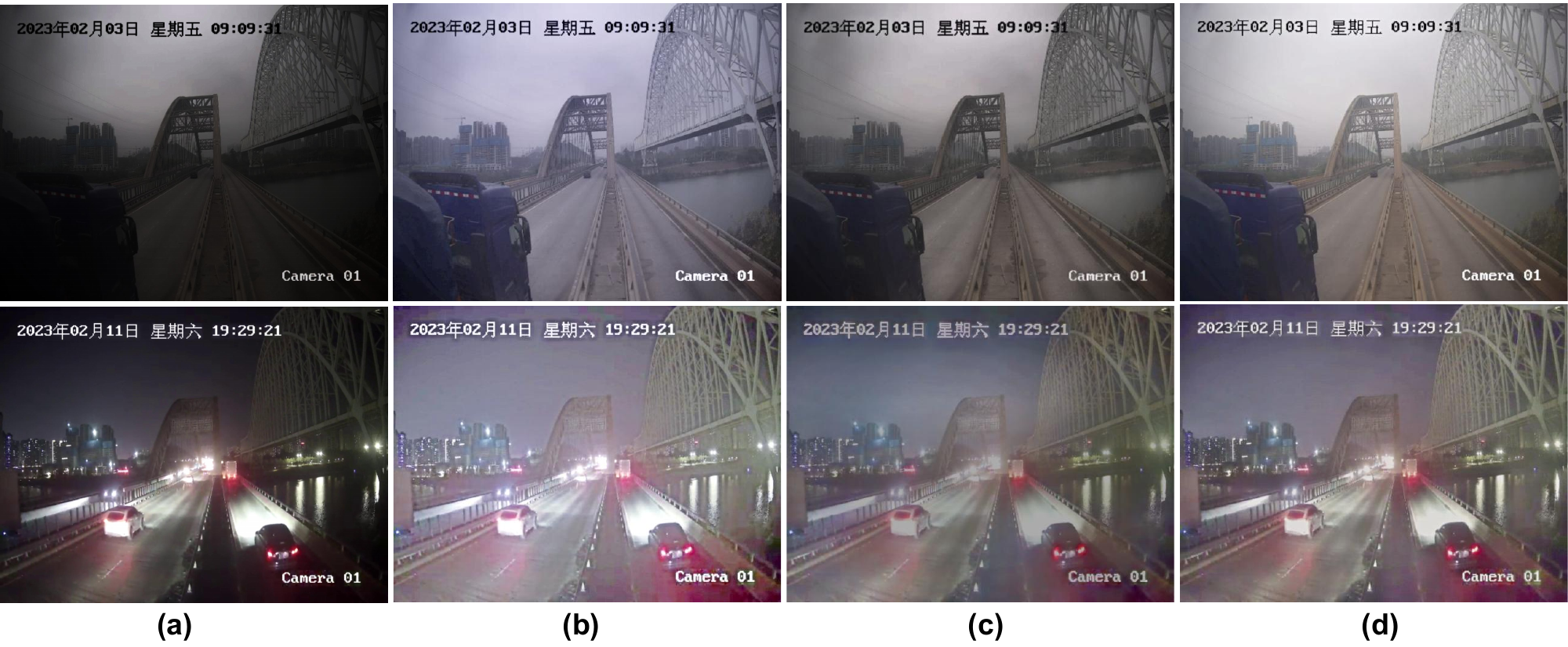}
    \caption{Visual detail comparison on the Traffic-297. (a) Input Image. (b) ZeroDCE++. (c) SNRNet. (d) ECAFormer. }
    \label{fig:visualResults2}
\end{figure}

\subsection{Experimental Configurations}
\subsubsection{Datasets} We conducted comparisons across multiple public datasets: LOL-v1 \cite{wei2018deep}, LOL-v2 \cite{yang2021sparse}, SID \cite{chen2019seeing} and SMID \cite{chen2018learning}. Additionally, we have collected the \textit{Traffic-297} dataset, which includes various traffic scenes. It contains an amount of real nighttime traffic scenes with complex lighting environments, which pose a significant challenge to low-light enhancement models.
\subsubsection{Explementation Details} We implemented the ECAFormer model using PyTorch. The model was trained on an NVIDIA RTX 4090 with 24 GB of VRAM using the Adam optimizer ($\beta_1$ = 0.9 and $\beta_2$ = 0.999), across a total of 250,000 iterations. The initial learning rate was set at 2 $\times$ 10$^{-4}$ and gradually reduced to 1 $\times$ 10$^{-6}$ through a cosine annealing schedule during the training process. Training samples were generated by randomly cropping 256 $\times$ 256 patches from pairs of low-/normal-light images. The batch size was set to 8. The training data was augmented with random rotations and flips to enhance variability and robustness. The training objective was to minimize both $\mathcal{L}_p$ \ref{perceptualloss} and $\mathcal{L}_c$ \ref{cbloss} between the enhanced images and their corresponding ground truths, ensuring high fidelity in the restoration of image details and color accuracy. To evaluate the performance, we employ the peak signal-to-noise ratio (PSNR) and structural similarity index (SSIM) \cite{1284395} as the primary metrics for evaluation.
% \begin{equation}
% \label{psnrformulation}
%     \text{PSNR} = 10 \cdot \log_{10} \left(\frac{\text{MAX}_I^2}{\text{MSE}}\right).
% \end{equation}
% \begin{equation}
% \label{ssimformulation}
%     \text{SSIM}(x, y) = \frac{(2\mu_x\mu_y + c_1)(2\sigma_{xy} + c_2)}{(\mu_x^2 + \mu_y^2 + c_1)(\sigma_x^2 + \sigma_y^2 + c_2)}.
% \end{equation}

\subsubsection{Comparison Methods}
To validate the effectiveness of our derived ECAFormer, we conducted extensive comparisons with several state-of-the-art conventional, two branch and guided  LLIE approaches. For the conventional LLIE, we compared the UFormer(UF) \cite{wang2022uformer}, EnlightenGAN(EnGAN) \cite{jiang2021enlightengan}, Restormer(RSTM) \cite{zamir2022restormer}, LLFormer(LLF) \cite{wang2023ultra}, and MIRNet \cite{zamir2020learning}.  For the two-branch LLIE, we compared the RetinexNet(RN) \cite{wei2018deep} and KinD \cite{zhang2019kindling}. For the guided LLIE, we compared the RUAS \cite{liu2021retinex}, DRBN \cite{yang2021band}, RetinexFormer(RF)\cite{cai2023retinexformer}, U2E-Net\cite{KHAN2024110490} and SNR-Net \cite{xu2022snr}

\subsection{Results Analysis}

\subsubsection{Quantitative Results} To validate the quantitative performance of ECAFormer, we compared our model across five public datasets using two metrics, PSNR and SSIM, with the results presented respectively in Table \ref{psnrcomparison} and Table \ref{ssimcomparison}. The results on our collected \textit{Traffic-297} dataset are shown in Table \ref{trafficresults}. Specifically, our ECAFormer achieves a 3.7\% improvement in PSNR and nearly a 1.3\% improvement in SSIM over the suboptimal method. The results indicate that our model achieves competitive performance while maintaining a relatively low parameter count and computational complexity.

\begin{table}[]
\caption{Comparison of SSIM of different methods across 5 public datasets.}
% \hspace{-1.8cm}
\centering
\resizebox{\textwidth}{!}{
\begin{tabular}{cccccc|cc}
\toprule
Methods & LOL-v1 & LOLv2-r         & LOLv2-s         & SID              & SMID             & AVG          & Params           \\
\midrule
RN(BMVC 2018)               & 0.560  & 0.567            & 0.798            & 0.578            & 0.684            & 0.6374           & 0.84             \\
KinD(ACMMM 2019)            & 0.790  & 0.641            & 0.578            & 0.583            & 0.634            & 0.6452           & 8.02             \\
EnGAN(TIP 2021)             & 0.650  & 0.617            & 0.734            & 0.543            & 0.674            & 0.6436           & 114.35           \\
RUAS(CVPR 2021)             & 0.720  & 0.723            & 0.652            & 0.581            & 0.744            & 0.6840           & 0.003            \\
DRBN(TIP 2021)              & 0.830  & 0.831            & 0.927            & 0.577            & 0.781            & 0.7892           & 5.27             \\
UF(CVPR 2022)               & 0.771  & 0.771            & 0.871            & 0.577            & 0.792            & 0.7564           & 5.29             \\
RSTM(CVPR 2022)             & 0.823  & 0.827            & 0.830            & 0.649            & 0.758            & 0.7774           & 26.13            \\
MIRNet(TPAMI 2022)          & 0.830  & 0.820            & 0.876            & 0.605            & 0.762            & 0.7786           & 31.76            \\
SNR(CVPR 2022)              & 0.842  & \underline{0.849}            & 0.928            & 0.625            & 0.805            & \underline{0.8098}           & 4.01             \\
LLF(AAAI 2023)              & 0.816  & 0.792            & 0.909            & \textbackslash{} & \textbackslash{} & \textbackslash{} & 24.55            \\
RF(ICCV 2023)    & 0.816  & 0.814            & \underline{0.930}            & \textbf{0.673}            & \textbf{0.815}            & 0.8096           & 1.61             \\
U2E-Net(PR 2024)            & \underline{0.847}  & \textbackslash{} & \textbackslash{} & \textbackslash{} & \textbackslash{} & \textbackslash{} & \textbackslash{} \\
\midrule
ECAFormer                   & \textbf{0.850}  & \textbf{0.853}            & \textbf{0.931}            & \underline{0.660}            & \underline{0.810}            & \textbf{0.8208}           & 2.50            \\
\bottomrule
\end{tabular}
}
\label{ssimcomparison}
\end{table}

\begin{table}[]
\caption{Comparison of PSNR and SSIM on the traffic-297 dataset.}
\setlength{\tabcolsep}{1mm}
\centering
\begin{tabular}{ccccccc}
\toprule
     & RN    & ZeroDCE++ & EnGAN & SNR   & RF    & ECAFormer \\
\midrule
PSNR & 18.56 & 18.35     & 18.19 & 30.06 & \underline{32.17} & \textbf{33.26}     \\
SSIM & 0.770 & 0.884     & 0.623 & 0.965 & \underline{0.975} & \textbf{0.978 }   \\
\bottomrule
\end{tabular}
\label{trafficresults}
\end{table}

\subsubsection{Quality Results} To validate the quality performance of ECAFormer, we present the visual comparison of the results, as shown in Figure \ref{fig:visualResults0}-\ref{fig:visualResults2}. As we can observe, the recent fully supervised learning method, SNRNet, did not retain sufficient detail and exhibited blurring. ZeroDCE++, which adjusts the pixel curves, introduced excessive noise. 
In the detailed comparison illustrated in Figure \ref{fig:visualResults1}, our method effectively preserves texture details while considering global brightness balance, indicating its ability to retain details and suppress noise. In Figure \ref{fig:visualResults3}, we showcase the performance of our model across different scenes, including color discrepancies caused by nighttime shooting. Specifically, our method was tested in complex scenarios from SDSD-in \cite{wang2021seeing}, SDSD-out, and SMID datasets, with results indicating strong performance of ECAFormer. In Figure \ref{fig:visualResults2}, we evaluated the performance in traffic scenes, specifically testing models trained on synthesized datasets under real-world nighttime scenarios. The complexity of lighting conditions in nighttime traffic scenes poses a significant challenge to models. While other methods may introduce noise and lose details, our model demonstrates exceptional visual performance by effectively preserving details while considering global illumination. In comparison, ZeroDCE++ exhibits overexposure issues, highlighting the superior visual results of our proposed model. These results indicate that our model adapts well to variations and produces high-quality outcomes closely approximating the ground truth, demonstrating the effectiveness of ECAFormer as proposed in this paper.
%在细节对比中，如Figure  \ref{fig:visualResults1}所示，我们的方法在考虑到了全局的亮度平衡的同时，良好的保留了纹理细节。 在Figure  \ref{fig:visualResults3}中，我们展示了我们的模型在不同的场景中的表现, 包括夜间拍摄导致的色差，结果表明我们的模型能够很好地适应。在Figure  \ref{visualResults2}中，我们测试了在交通场景下的表现。其中，夜间交通场景包含着复杂的光源，对低光增强模型有着较大考验，例如ZeroDCE++出现了过曝的现象。相比之下我们的模型能够在考虑到全局光照的同时，良好的保留细节，并展现出最好的视觉效果。

\begin{figure}[t]
    \centering
\includegraphics[width=\textwidth]{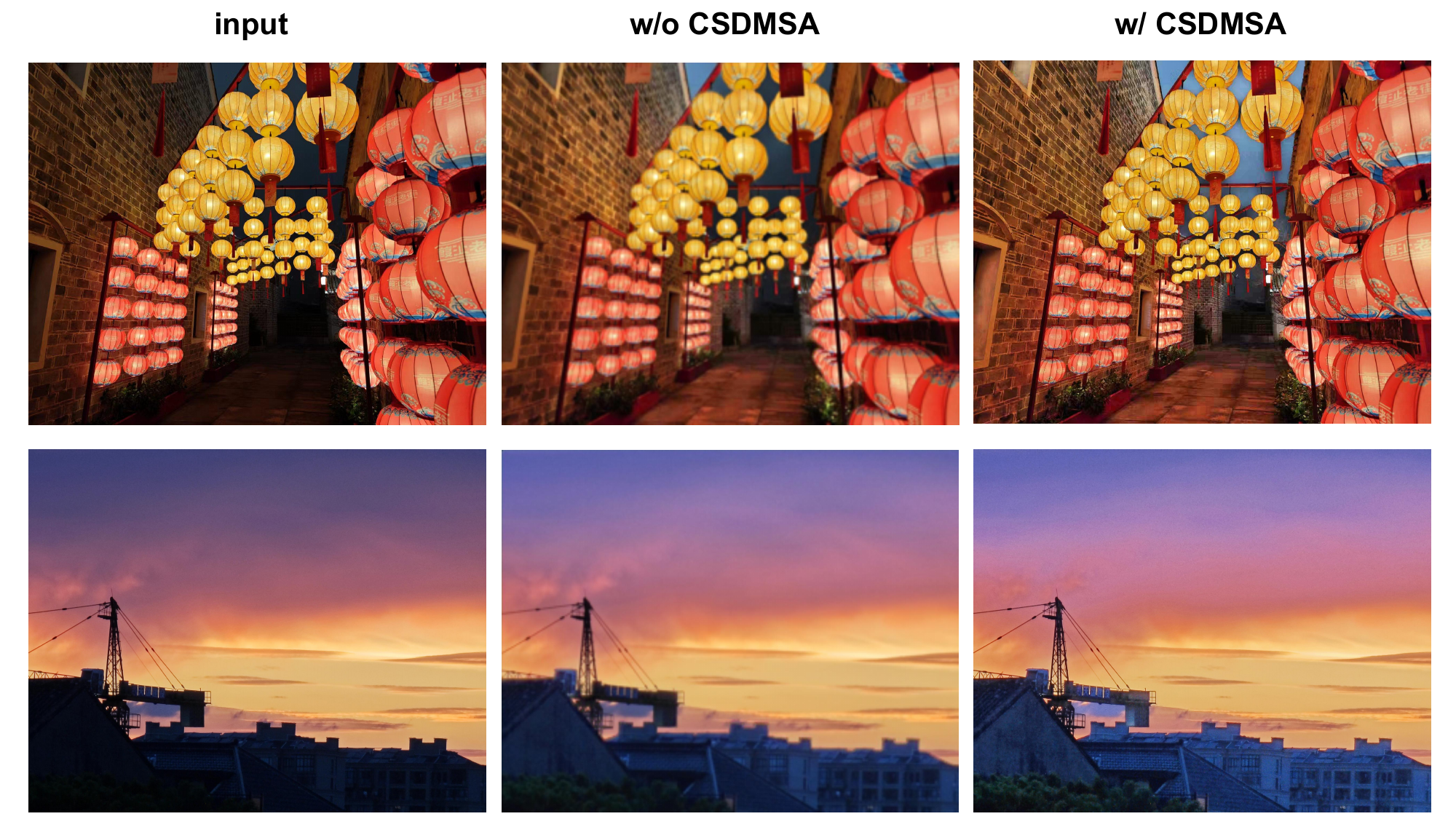}
    \caption{The ablation experiment of CSDMSA. Without the CSDMSA module, the images become blurred.}
    \label{fig:CSDMSA}
    % TODO
\end{figure}

%同时，我们还通过DMSA来处理跨层的残差信息来减少下采样带来的细节丢失。我们验证了DMSA模组对残差信息的融合能力。通过Figure  \ref{CSDMSA}可以看出，在没有CSDMSA模块的情况下，增强之后的图片会变得模糊不清，缺少细节信息。因此可以说明CSDMSA能够有效利用跨层的残差信息，来抑制下采样导致的细节的丢失
\begin{figure}[t]
    \centering
\includegraphics[width=\textwidth]{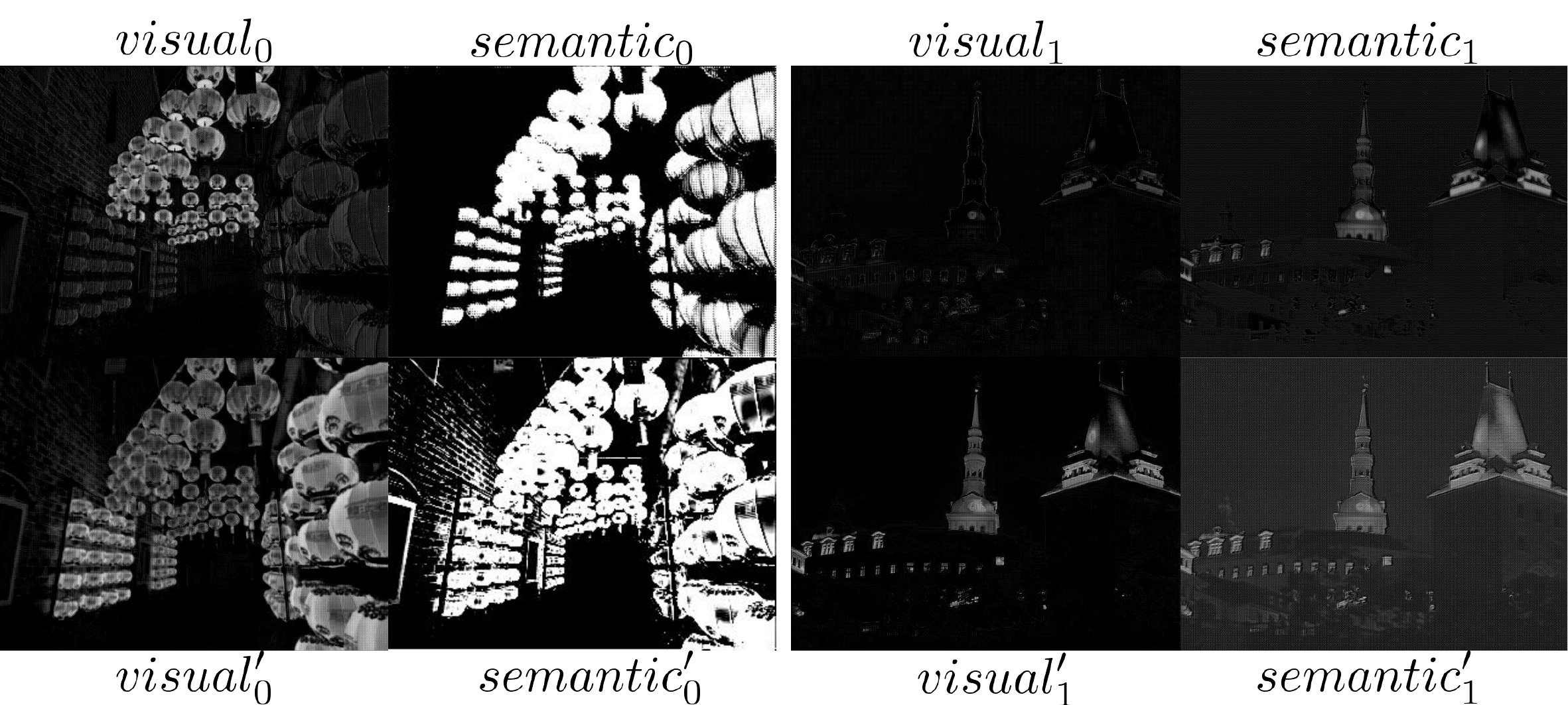}
    \caption{Visualization of the two features before and after interaction through DMSA. The first row shows before interaction, and the second row shows after interaction.}
    \label{fig:interlayer}
    % TODO
\end{figure}

\subsection{Ablation Study}
To assess the efficacy of the different modules, including (1) Visual-Semantic Feature (VSF): remove the visual semantic module and directly pass the input to DMSA., (2) Loss: only use $L_{1}$ loss without perceptual loss and charbonnier loss, and (3) DMSA: replace DMSA with conventional MHSA, we made adjustments to the model structure and presented the PSNR results in given Table \ref{ablationresults} and Figures \ref{fig:CSDMSA} - \ref{fig:interlayer}. The results presented in Table \ref{ablationresults} suggest that all three components positively impact the model's performance and are effective. Among them, the DMSA module shows the most significant improvement in feature fusion.

%\textbf{1) Visual-Semantic Feature(VSF).}
%We validated the effectiveness of the shallow-deep convolution module. When this module is not utilized, the input image is directly reshaped to the format required by the DMSA for input.

%\textbf{2) Loss.}
%Verified the effectiveness of the loss functions by comparing the Mean Squared Error (MSE) loss with the two types of losses used in this study.

%\textbf{3) DMSA.}
%Replaced the DMSA with a classic self-attention mechanism to validate the performance of this module in feature fusion.
\begin{table}[t]
\centering
\caption{Ablation study results on SDSD-out.}
\begin{tabular}{cccc}
\toprule
VSF & Loss & DMSA & PSNR(avg)  \\ \midrule
          & \checkmark           & \checkmark       & 28.56 \\
\checkmark         &             & \checkmark       & 29.41 \\
\checkmark         & \checkmark           &         & 28.38 \\
\checkmark         & \checkmark           & \checkmark       & 29.57 \\ \bottomrule
\end{tabular}
\label{ablationresults}
\end{table}

Furthermore, in ECAFormer, we utilize the DMSA module to effectively manage cross-layer residual information, thereby minimizing the loss of details resulting from downsampling. This validates the capability of the CSDMSA module to integrate residual information, as depicted in Figure \ref{fig:CSDMSA}. It is evident that without the CSDMSA module, the enhanced images suffer from blurring and significant loss of image detail. This outcome indicates that the CSDMSA module effectively mitigates the loss of details caused by downsampling by utilizing cross-layer residual information.

Besides, as depicted in Figure \ref{fig:interlayer}, DMSA effectively facilitates the seamless integration of visual and semantic features. The findings demonstrate that visual features complement semantic features by providing additional intricate details, while semantic features offer long-range dependencies for visual features, resulting in improved brightness uniformity. Therefore, our designed DSMA significantly enhances the complementary information of visual and semantic features.
%如图 \ref{fig:interlayer}所示，DMSA能够有效的相互融合视觉特征与语义特征。 视觉特征能够为语义特征补充细节信息。同时，语义特征能够为视觉特征提供长程依赖，从而有更好的亮度统一性。
\section{Conclusion}
\label{Conclusion}
In this paper, we propose a transformer-based method for LLIE. Specifically, we use DMSA blocks to enhance low-light images by enhancing the semantic and visual components of the image. Furthermore, the ECAFormer network combines the advantages of local detailed feature information extraction in CNNs and global information extraction in transformer networks, which effectively reveals the local and global information of the image. Thus, the detailed information can be preserved after enhancement. Extensive experimental results validate the competitive advantages of ECAFormer compared to other state-of-the-art algorithms. In our future research, we will focus on refining the training approach to enable the model to effectively utilize disparate datasets and integrate additional prior knowledge to enhance its feature extraction capabilities. 
\section*{ACKNOWLEDGMENTS}

This work was partially supported by the National Natural Science Foundation of China (62306051,6230073196,62481540175,), the Fundamental Research Funds for the Central Universities (NJ2022028), and the Scientific and the Technological Research Program of Chongqing Municipal Education Commission (KJQN202300718). Weikai Li was partly supported by the  National Natural Science Foundation of China (62076124).

% \textbf{Future work.} In recent years, models trained on extensive datasets have progressively evolved. The integration of prior knowledge from large models can mitigate the limitations of current loss functions and extend to unsupervised learning. The two-dimensional Fourier transform holds substantial potential in handling image details and could be incorporated into the DMSA module proposed in this paper, enabling interactions in both the frequency and time domains.

 \bibliographystyle{elsarticle-num} 
 \bibliography{mybibfile}

\begin{thebibliography}{10}
\expandafter\ifx\csname url\endcsname\relax
  \def\url#1{\texttt{#1}}\fi
\expandafter\ifx\csname urlprefix\endcsname\relax\def\urlprefix{URL }\fi
\expandafter\ifx\csname href\endcsname\relax
  \def\href#1#2{#2} \def\path#1{#1}\fi

\bibitem{ranft2016role}
B.~Ranft, C.~Stiller, The role of machine vision for intelligent vehicles, IEEE Transactions on Intelligent vehicles 1~(1) (2016) 8--19.

\bibitem{cai2020vtgnet}
P.~Cai, Y.~Sun, H.~Wang, M.~Liu, Vtgnet: A vision-based trajectory generation network for autonomous vehicles in urban environments, IEEE Transactions on Intelligent Vehicles 6~(3) (2020) 419--429.

\bibitem{bresson2017simultaneous}
G.~Bresson, Z.~Alsayed, L.~Yu, S.~Glaser, Simultaneous localization and mapping: A survey of current trends in autonomous driving, IEEE Transactions on Intelligent Vehicles 2~(3) (2017) 194--220.

\bibitem{fu2021let}
L.~Fu, H.~Yu, F.~Juefei-Xu, J.~Li, Q.~Guo, S.~Wang, Let there be light: Improved traffic surveillance via detail preserving night-to-day transfer, IEEE Transactions on Circuits and Systems for Video Technology 32~(12) (2021) 8217--8226.

\bibitem{BULUSWAR1998245}
S.~D. Buluswar, B.~A. Draper, \href{https://www.sciencedirect.com/science/article/pii/S0952197697000791}{Color machine vision for autonomous vehicles}, Engineering Applications of Artificial Intelligence 11~(2) (1998) 245--256.
\newblock \href {https://doi.org/https://doi.org/10.1016/S0952-1976(97)00079-1} {\path{doi:https://doi.org/10.1016/S0952-1976(97)00079-1}}.
\newline\urlprefix\url{https://www.sciencedirect.com/science/article/pii/S0952197697000791}

\bibitem{lore2017llnet}
K.~G. Lore, A.~Akintayo, S.~Sarkar, Llnet: A deep autoencoder approach to natural low-light image enhancement, Pattern Recognition 61 (2017) 650--662.

\bibitem{wang2022uformer}
Z.~Wang, X.~Cun, J.~Bao, W.~Zhou, J.~Liu, H.~Li, Uformer: A general u-shaped transformer for image restoration, in: Proceedings of the IEEE/CVF conference on computer vision and pattern recognition, 2022, pp. 17683--17693.

\bibitem{ZHANG2023106972}
D.~Zhang, Z.~He, X.~Zhang, Z.~Wang, W.~Ge, T.~Shi, Y.~Lin, \href{https://www.sciencedirect.com/science/article/pii/S0952197623011569}{Underwater image enhancement via multi-scale fusion and adaptive color-gamma correction in low-light conditions}, Engineering Applications of Artificial Intelligence 126 (2023) 106972.
\newblock \href {https://doi.org/https://doi.org/10.1016/j.engappai.2023.106972} {\path{doi:https://doi.org/10.1016/j.engappai.2023.106972}}.
\newline\urlprefix\url{https://www.sciencedirect.com/science/article/pii/S0952197623011569}

\bibitem{ZHANG2024107793}
Y.~Zhang, W.~Xu, C.~Lyu, \href{https://www.sciencedirect.com/science/article/pii/S0952197623019772}{Semantic-aware enhancement: Integrating semantic compensation with 3-dimensional lookup tables for low-light image enhancement}, Engineering Applications of Artificial Intelligence 130 (2024) 107793.
\newblock \href {https://doi.org/https://doi.org/10.1016/j.engappai.2023.107793} {\path{doi:https://doi.org/10.1016/j.engappai.2023.107793}}.
\newline\urlprefix\url{https://www.sciencedirect.com/science/article/pii/S0952197623019772}

\bibitem{ZHUANG2021104171}
P.~Zhuang, C.~Li, J.~Wu, \href{https://www.sciencedirect.com/science/article/pii/S095219762100018X}{Bayesian retinex underwater image enhancement}, Engineering Applications of Artificial Intelligence 101 (2021) 104171.
\newblock \href {https://doi.org/https://doi.org/10.1016/j.engappai.2021.104171} {\path{doi:https://doi.org/10.1016/j.engappai.2021.104171}}.
\newline\urlprefix\url{https://www.sciencedirect.com/science/article/pii/S095219762100018X}

\bibitem{LI2023106457}
D.~Li, J.~Zhou, S.~Wang, D.~Zhang, W.~Zhang, R.~Alwadai, F.~Alenezi, P.~Tiwari, T.~Shi, \href{https://www.sciencedirect.com/science/article/pii/S0952197623006413}{Adaptive weighted multiscale retinex for underwater image enhancement}, Engineering Applications of Artificial Intelligence 123 (2023) 106457.
\newblock \href {https://doi.org/https://doi.org/10.1016/j.engappai.2023.106457} {\path{doi:https://doi.org/10.1016/j.engappai.2023.106457}}.
\newline\urlprefix\url{https://www.sciencedirect.com/science/article/pii/S0952197623006413}

\bibitem{land1971lightness}
E.~H. Land, J.~J. McCann, Lightness and retinex theory, Josa 61~(1) (1971) 1--11.

\bibitem{ma2021learning}
L.~Ma, R.~Liu, J.~Zhang, X.~Fan, Z.~Luo, Learning deep context-sensitive decomposition for low-light image enhancement, IEEE Transactions on Neural Networks and Learning Systems 33~(10) (2021) 5666--5680.

\bibitem{tang2022drlie}
L.~Tang, J.~Ma, H.~Zhang, X.~Guo, Drlie: Flexible low-light image enhancement via disentangled representations, IEEE transactions on neural networks and learning systems 35~(2) (2022) 2694--2707.

\bibitem{AHERRAHROU2023109643}
N.~Aherrahrou, H.~Tairi, \href{https://www.sciencedirect.com/science/article/pii/S0031320323003448}{A novel cancelable finger vein templates based on ldm and retinexgan}, Pattern Recognition 142 (2023) 109643.
\newblock \href {https://doi.org/https://doi.org/10.1016/j.patcog.2023.109643} {\path{doi:https://doi.org/10.1016/j.patcog.2023.109643}}.
\newline\urlprefix\url{https://www.sciencedirect.com/science/article/pii/S0031320323003448}

\bibitem{HE2023106969}
X.~He, Z.~Chen, L.~Dai, L.~Liang, J.~Wu, B.~Sheng, \href{https://www.sciencedirect.com/science/article/pii/S0952197623011533}{Global-and-local aware network for low-light image enhancement}, Engineering Applications of Artificial Intelligence 126 (2023) 106969.
\newblock \href {https://doi.org/https://doi.org/10.1016/j.engappai.2023.106969} {\path{doi:https://doi.org/10.1016/j.engappai.2023.106969}}.
\newline\urlprefix\url{https://www.sciencedirect.com/science/article/pii/S0952197623011533}

\bibitem{li2021low}
C.~Li, C.~Guo, L.~Han, J.~Jiang, M.-M. Cheng, J.~Gu, C.~C. Loy, Low-light image and video enhancement using deep learning: A survey, IEEE transactions on pattern analysis and machine intelligence 44~(12) (2021) 9396--9416.

\bibitem{LI2025109749}
X.~Li, W.~Wang, Y.~Han, X.~Feng, \href{https://www.sciencedirect.com/science/article/pii/S0952197624019080}{Structure aware transfer function network for low light image enhancement}, Engineering Applications of Artificial Intelligence 141 (2025) 109749.
\newblock \href {https://doi.org/https://doi.org/10.1016/j.engappai.2024.109749} {\path{doi:https://doi.org/10.1016/j.engappai.2024.109749}}.
\newline\urlprefix\url{https://www.sciencedirect.com/science/article/pii/S0952197624019080}

\bibitem{wei2018deep}
C.~Wei, W.~Wang, W.~Yang, J.~Liu, Deep retinex decomposition for low-light enhancement, arXiv preprint arXiv:1808.04560 (2018).

\bibitem{liu2021retinex}
R.~Liu, L.~Ma, J.~Zhang, X.~Fan, Z.~Luo, Retinex-inspired unrolling with cooperative prior architecture search for low-light image enhancement, in: Proceedings of the IEEE/CVF conference on computer vision and pattern recognition, 2021, pp. 10561--10570.

\bibitem{jiang2021enlightengan}
Y.~Jiang, X.~Gong, D.~Liu, Y.~Cheng, C.~Fang, X.~Shen, J.~Yang, P.~Zhou, Z.~Wang, Enlightengan: Deep light enhancement without paired supervision, IEEE transactions on image processing 30 (2021) 2340--2349.

\bibitem{zhou2022lednet}
S.~Zhou, C.~Li, C.~C. Loy, Lednet: Joint low-light enhancement and deblurring in the dark (2022).
\newblock \href {http://arxiv.org/abs/2202.03373} {\path{arXiv:2202.03373}}.

\bibitem{FAN2023105632}
S.~Fan, W.~Liang, D.~Ding, H.~Yu, \href{https://www.sciencedirect.com/science/article/pii/S0952197622006224}{Lacn: A lightweight attention-guided convnext network for low-light image enhancement}, Engineering Applications of Artificial Intelligence 117 (2023) 105632.
\newblock \href {https://doi.org/https://doi.org/10.1016/j.engappai.2022.105632} {\path{doi:https://doi.org/10.1016/j.engappai.2022.105632}}.
\newline\urlprefix\url{https://www.sciencedirect.com/science/article/pii/S0952197622006224}

\bibitem{LIU2024109012}
J.~Liu, S.~Wang, C.~Chen, Q.~Hou, \href{https://www.sciencedirect.com/science/article/pii/S0952197624011709}{Dfp-net: An unsupervised dual-branch frequency-domain processing framework for single image dehazing}, Engineering Applications of Artificial Intelligence 136 (2024) 109012.
\newblock \href {https://doi.org/https://doi.org/10.1016/j.engappai.2024.109012} {\path{doi:https://doi.org/10.1016/j.engappai.2024.109012}}.
\newline\urlprefix\url{https://www.sciencedirect.com/science/article/pii/S0952197624011709}

\bibitem{ZHU2023106866}
P.~Zhu, Y.~Liu, Y.~Wen, M.~Xu, X.~Fu, S.~Liu, \href{https://www.sciencedirect.com/science/article/pii/S0952197623010503}{Unsupervised underwater image enhancement via content-style representation disentanglement}, Engineering Applications of Artificial Intelligence 126 (2023) 106866.
\newblock \href {https://doi.org/https://doi.org/10.1016/j.engappai.2023.106866} {\path{doi:https://doi.org/10.1016/j.engappai.2023.106866}}.
\newline\urlprefix\url{https://www.sciencedirect.com/science/article/pii/S0952197623010503}

\bibitem{CHEN2024109207}
X.~Chen, X.~Zhou, W.~Sun, Y.~Zhang, \href{https://www.sciencedirect.com/science/article/pii/S0952197624013654}{Image deraining via invertible disentangled representations}, Engineering Applications of Artificial Intelligence 137 (2024) 109207.
\newblock \href {https://doi.org/https://doi.org/10.1016/j.engappai.2024.109207} {\path{doi:https://doi.org/10.1016/j.engappai.2024.109207}}.
\newline\urlprefix\url{https://www.sciencedirect.com/science/article/pii/S0952197624013654}

\bibitem{xu2022snr}
X.~Xu, R.~Wang, C.-W. Fu, J.~Jia, Snr-aware low-light image enhancement, in: Proceedings of the IEEE/CVF conference on computer vision and pattern recognition, 2022, pp. 17714--17724.

\bibitem{cai2023retinexformer}
Y.~Cai, H.~Bian, J.~Lin, H.~Wang, R.~Timofte, Y.~Zhang, Retinexformer: One-stage retinex-based transformer for low-light image enhancement, in: Proceedings of the IEEE/CVF International Conference on Computer Vision, 2023, pp. 12504--12513.

\bibitem{johnson2016perceptual}
J.~Johnson, A.~Alahi, L.~Fei-Fei, Perceptual losses for real-time style transfer and super-resolution, in: Computer Vision--ECCV 2016: 14th European Conference, Amsterdam, The Netherlands, October 11-14, 2016, Proceedings, Part II 14, Springer, 2016, pp. 694--711.

\bibitem{yang2021sparse}
W.~Yang, W.~Wang, H.~Huang, S.~Wang, J.~Liu, Sparse gradient regularized deep retinex network for robust low-light image enhancement, IEEE Transactions on Image Processing 30 (2021) 2072--2086.

\bibitem{chen2019seeing}
C.~Chen, Q.~Chen, M.~N. Do, V.~Koltun, Seeing motion in the dark, in: Proceedings of the IEEE/CVF International conference on computer vision, 2019, pp. 3185--3194.

\bibitem{chen2018learning}
C.~Chen, Q.~Chen, J.~Xu, V.~Koltun, Learning to see in the dark, in: Proceedings of the IEEE conference on computer vision and pattern recognition, 2018, pp. 3291--3300.

\bibitem{1284395}
Z.~Wang, A.~Bovik, H.~Sheikh, E.~Simoncelli, Image quality assessment: from error visibility to structural similarity, IEEE Transactions on Image Processing 13~(4) (2004) 600--612.
\newblock \href {https://doi.org/10.1109/TIP.2003.819861} {\path{doi:10.1109/TIP.2003.819861}}.

\bibitem{zamir2022restormer}
S.~W. Zamir, A.~Arora, S.~Khan, M.~Hayat, F.~S. Khan, M.-H. Yang, Restormer: Efficient transformer for high-resolution image restoration, in: Proceedings of the IEEE/CVF conference on computer vision and pattern recognition, 2022, pp. 5728--5739.

\bibitem{wang2023ultra}
T.~Wang, K.~Zhang, T.~Shen, W.~Luo, B.~Stenger, T.~Lu, Ultra-high-definition low-light image enhancement: A benchmark and transformer-based method, in: Proceedings of the AAAI Conference on Artificial Intelligence, Vol.~37, 2023, pp. 2654--2662.

\bibitem{zamir2020learning}
S.~W. Zamir, A.~Arora, S.~Khan, M.~Hayat, F.~S. Khan, M.-H. Yang, L.~Shao, Learning enriched features for real image restoration and enhancement, in: Computer Vision--ECCV 2020: 16th European Conference, Glasgow, UK, August 23--28, 2020, Proceedings, Part XXV 16, Springer, 2020, pp. 492--511.

\bibitem{zhang2019kindling}
Y.~Zhang, J.~Zhang, X.~Guo, Kindling the darkness: A practical low-light image enhancer, in: Proceedings of the 27th ACM international conference on multimedia, 2019, pp. 1632--1640.

\bibitem{yang2021band}
W.~Yang, S.~Wang, Y.~Fang, Y.~Wang, J.~Liu, Band representation-based semi-supervised low-light image enhancement: Bridging the gap between signal fidelity and perceptual quality, IEEE Transactions on Image Processing 30 (2021) 3461--3473.

\bibitem{KHAN2024110490}
R.~Khan, A.~Mehmood, F.~Shahid, Z.~Zheng, M.~M. Ibrahim, \href{https://www.sciencedirect.com/science/article/pii/S0031320324002413}{Lit me up: A reference free adaptive low light image enhancement for in-the-wild conditions}, Pattern Recognition 153 (2024) 110490.
\newblock \href {https://doi.org/https://doi.org/10.1016/j.patcog.2024.110490} {\path{doi:https://doi.org/10.1016/j.patcog.2024.110490}}.
\newline\urlprefix\url{https://www.sciencedirect.com/science/article/pii/S0031320324002413}

\bibitem{wang2021seeing}
R.~Wang, X.~Xu, C.-W. Fu, J.~Lu, B.~Yu, J.~Jia, Seeing dynamic scene in the dark: A high-quality video dataset with mechatronic alignment, in: Proceedings of the IEEE/CVF International Conference on Computer Vision, 2021, pp. 9700--9709.

\end{thebibliography}



% %% For numbered reference style

%   \textit{\LaTeX: a document preparation system},
%   Addison Wesley, Massachusetts,

% \end{thebibliography}
\end{document}